\documentclass[conference]{IEEEtran}
\IEEEoverridecommandlockouts
\usepackage{cite}
\usepackage{amsmath,amssymb,amsfonts}
\usepackage{graphicx}
\usepackage{textcomp}
\usepackage{xcolor}
\usepackage{graphicx}
\usepackage{textcomp}
\usepackage{xcolor}
\def\BibTeX{{\rm B\kern-.05em{\sc i\kern-.025em b}\kern-.08em
    T\kern-.1667em\lower.7ex\hbox{E}\kern-.125emX}}
\usepackage{floatrow}

\usepackage[utf8]{inputenc}
\usepackage{graphicx}
\usepackage{dutchcal}
\usepackage[mathscr]{eucal}
\usepackage{commath}
\usepackage{latexsym}
\usepackage{caption}

\usepackage{setspace}
\usepackage[english]{babel}
\usepackage[noend]{algpseudocode}
\usepackage{algorithmicx}
\usepackage{booktabs}
\usepackage{algorithm}
\usepackage[utf8]{inputenc}
\usepackage{nicefrac}
\usepackage{mathtools}
\usepackage{centernot}
\usepackage{makecell}
\usepackage{tabularx}
\usepackage{xcolor}
\usepackage{mathtools}
\usepackage{soul}
\usepackage{svg}
\usepackage{hyperref}
\usepackage{comment}
\usepackage{dsfont}
\usepackage{booktabs}
\usepackage{listings}% http://ctan.org/pkg/listings
\usepackage{enumitem}

\newfloatcommand{capbtabbox}{table}[][\FBwidth]

\newcommand{\name}{BiasBuster}
\newtheorem{theorem}{Theorem}[section]
\newtheorem{lemma}[theorem]{Lemma}

\DeclareMathAlphabet{\mathcal}{OMS}{cmsy}{m}{n}
\SetMathAlphabet{\mathcal}{bold}{OMS}{cmsy}{b}{n}
\def\BibTeX{{\rm B\kern-.05em{\sc i\kern-.025em b}\kern-.08em
    T\kern-.1667em\lower.7ex\hbox{E}\kern-.125emX}}
\begin{document}

\title{\name{}: a Neural Approach for Accurate Estimation of Population Statistics using Biased Location Data}

\author{\IEEEauthorblockN{Sepanta Zeighami}
\IEEEauthorblockA{\textit{University of Southern California}\\zeighami@usc.edu}
\and
\IEEEauthorblockN{Cyrus Shahabi}
\IEEEauthorblockA{\textit{University of Southern California}\\shahabi@usc.edu}

}
\pagestyle{plain}

\setlength{\abovedisplayskip}{0pt}
\setlength{\belowdisplayskip}{0pt}
\setlength{\abovedisplayshortskip}{0pt}
\setlength{\belowdisplayshortskip}{0pt}

\setlength{\textfloatsep}{0pt}
\setlength{\floatsep}{0pt}
\setlength{\intextsep}{0pt}
\setlength{\dbltextfloatsep}{0pt}
\setlength{\dblfloatsep}{0pt}
\setlength{\abovecaptionskip}{0pt}
\setlength{\belowcaptionskip}{0pt}

\maketitle

\begin{abstract}
While extremely useful (e.g., for COVID-19 forecasting and policy-making, urban mobility analysis and marketing, and obtaining business insights), location data collected from mobile devices often contain data from a biased population subset, with some communities over or underrepresented in the collected datasets. As a result, aggregate statistics calculated from such datasets (as is done by various companies including Safegraph, Google, and Facebook), while ignoring the bias, leads to an inaccurate representation of population statistics. Such statistics will not only be generally inaccurate, but the error will disproportionately impact different population subgroups (e.g., because they ignore the underrepresented communities). This has dire consequences, as these datasets are used for sensitive decision-making such as COVID-19 policymaking.  This paper tackles the problem of providing accurate population statistics using such biased datasets. We show that statistical debiasing, although in some cases useful, often fails to improve accuracy. We then propose BiasBuster, a neural network approach that utilizes the correlations between population statistics and location characteristics to provide accurate estimates of population statistics. Extensive experiments on real-world data show that BiasBuster improves accuracy by up to 2 times in general and up to 3 times for underrepresented populations.
\end{abstract}

\if 0
\fi

\section{Introduction}
Location data collected from mobile devices is useful: \cite{chang2021mobility,zeighami2021estimating,rambhatla2022toward,hu2021human} use such datasets for COVID-19 forecasting and policy-making, \cite{haraguchi2022human, song2022factors, zhao2016urban} use them for urban mobility analysis  and \cite{sun2022predicting,Federal2023CARTS,jin2022customer} use them for marketing and obtaining business insights. A concrete example is deciding closure policies for COVID, which depends on how many people go to a location and how long they stay there (a place should be closed if many people stay there for a sufficiently long duration). Such studies use aggregate statistics obtained from the observed data. The observed data is a set of location signals (e.g., GPS pings from cellphones) from which aggregate statistics are calculated (e.g., number of people in an area, average time people stay at a location and average distance they travel to get there). 

\begin{table}[t]
    \centering
\begin{tabular}{c|c|c|c|c}
\textbf{City}&\textbf{Adult}&\textbf{Senior}&\textbf{Child}&\textbf{Median Income}\\\hline
Houston&0.31 &-0.12 &-0.24 &-0.08 \\\hline
%Los Angeles&0.03 &0.01 &-0.05 &-0.10 \\\hline
Chicago&0.09 &-0.02 &-0.19 &-0.18 \\\hline
%Salt Lake&0.49 &-0.03 &-0.34 &-0.33 \\\hline
San Francisco&0.16 &0.04 &-0.30 &-0.24 \\\hline
%Boston&0.16 &0.18 &-0.22 &0.06 \\\hline
%Kansas City&0.40 &0.05 &-0.37 &-0.12 \\\hline
%Phoenix&0.38 &-0.06 &-0.24 &-0.23 \\\hline
%Milwaukee&0.17 &-0.08 &-0.16 &-0.35 \\\hline
Tulsa&0.41 &-0.12 &-0.22 &-0.29 \\\hline
Fargo&0.63 &-0.38 &-0.40 &-0.39 
\end{tabular}
    \caption{Correlation between observed proportion of neighbourhood population and neighbourhood demographics}
    \label{tab:sampling_bias}
\end{table}

However, observed location data is often a biased subset of the population data. More data is available for some sub-groups of the population, a bias that occurs due to the means of data collection. Location datasets are collected from mobile apps, which are often used by different sub-groups within a population at varying degrees. Android phones and iPhones are used by different demographics \cite{iphonevsandroidb, iphonevsandroida}, and dependence on demographics such as age is more broadly true across different app users \cite{appdemb, appdema}. To quantify such biases, we analyzed the location dataset used by Safegraph, a popular data curator, which provides aggregate location-based statistics \cite{sgp} extensively used for COVID studies \cite{chang2021mobility,zeighami2021estimating,rambhatla2022toward,hu2021human} and other applications \cite{haraguchi2022human, song2022factors, zhao2016urban, sun2022predicting,Federal2023CARTS,jin2022customer}%,ballantyne2022framework,safegraph2023clear}. 
%, releases aggregate statistics mentioned above from such biased datasets. By analyzing the underlying location dataset based on which Safegraph releases aggregate statistics (provided by Veraset) XX, 
We observe that the dataset contains more data for the adult and low-income populations\footnote{We suspect this is the case because the data is mostly collected from Android phones.} while it contains less data for the senior or child populations. This is summarized in Table~\ref{tab:sampling_bias}, which shows the Pearson correlation between the portion of the neighbourhood's population for which data was available and various neighbourhood demographics (see Sec.~\ref{sec:exp} for methodology and details). A higher correlation for an attribute (e.g., income) means neighbourhoods with a higher value of that attribute had a larger portion of their population in the observed data. For instance, in Fargo, we see a relatively strong negative correlation between income and the population's representation in the observed GPS data.

The problem, then, is to provide accurate aggregate population statistics while having access only to a biased sample of the population's location. An approach that reports aggregate statistics but is oblivious to the present bias leads to inaccurate estimates. In this case, the estimation error is due to using a \textit{biased estimator} of the population statistic. Consequently, we not only obtain inaccurate estimations in general, but the estimation error will also disproportionately impact different population subgroups. For instance, less data for the senior population leads to larger errors for such a population, while more data for people with low income can lead to an overestimation of densities in low-income neighbourhoods (because low-income neighbourhoods will appear to be more densely populated compared with other neighoubhoods in our observed data). This can, for example, result in COVID-19 policies that put undue burdens on such (often more vulnerable) populations.  It is, therefore, essential to provide accurate estimates for all population subgroups.

To improve the accuracy, one approach is to use statistical debiasing to provide an \textit{unbiased estimator} of the population statistics (i.e., an estimator equal to the population statistic on expectation). This is achieved by utilizing the probabilities of users being sampled, and weighting the samples according to those probabilities when estimating the population statistics. Although such an approach eliminates the bias in estimation, the variance of the estimator (due to the randomness in sampling) leads to errors in the estimations. We observed that debiasing helped improve the accuracy when calculating \texttt{COUNT} statistics (e.g., number of users at a location), but led to worse estimates for \texttt{AVG} statistics (e.g., average duration people stay at a location). Indeed, in our datasets, we observed the variance of attributes such as visit duration and distance traveled were much larger than the variance of the number of visits. For example, many users have a similar number of visits to a location (e.g., most people go to the same restaurant at most once a week), while there are variations across how long people stay there (e.g., short stays to pick-up food, longer stay to eat and even longer stays for social gatherings). This leads to larger variance when estimating \texttt{AVG} queries, which is further amplified by debiasing as often samples needed to be weighted by large values. As a result, we observe an example of the bias/variance trade-off, where debiasing eliminates the bias but the increased variance of the estimator leads to a larger error for \texttt{AVG} statistics.  

%This is due to the larger variance of the attributes when answering AVG queries. As an example, consider two users who visit a restaurant, where the first user only has one 60m visit while the second user has two visits, 5m and 10m each (e.g., second user orders pick-up while the first user dines in). Randomly sampling one user (even in the unbiased case) to estimate total number of visits to the restaurant, we obtain either 2 or 4 estimates with average relative error of 1/3. However, to estimatie visit duration, we obtain estimates of either X and X with average error of X. Overall, in the case of average, we observe the bias/variance trade-off, where reducing bias through statistical debiasing increases variance, and often leads to worse answers.  

\begin{figure}
    \centering
    \includegraphics[width=\columnwidth]{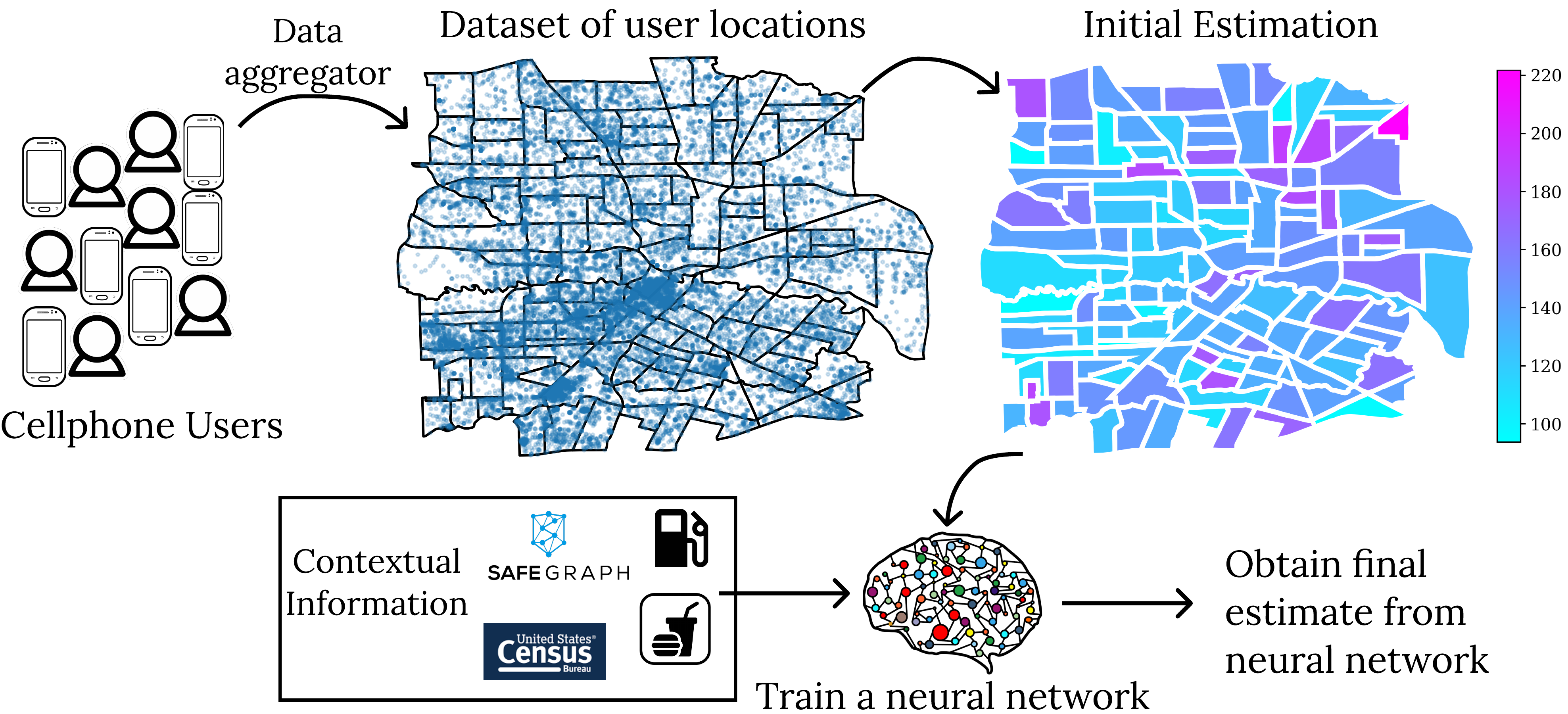}
    \caption{\name{} end-to-end pipeline}
    \label{fig:pipeline}
\end{figure}

In this paper, we introduce \name{}, a learned estimator that utilizes patterns in the aggregate statistics to provide accurate estimates of population statistics for all subgroups of the population. The end-to-end pipeline is presented in Fig.~\ref{fig:pipeline}. Given a set of location data collected from mobile users, \name{} first uses statistical methods described above to obtain initial estimates for population statistics. Then, it uses these initial estimates, together with contextual information available about different locations to train a model that learns the underlying correlations between location characteristics and the population statistics (e.g., people tend to stay shorter in gas stations than in restaurants). Learning such correlations leads to a model that provides accurate estimation for the population statistics.  We experimentally verify this observation, showing that \name{} reduces the estimation error by a factor of 2 in general, and specifically reduces the error by a factor of 3 for under-sampled neighbourhoods.

\name{} uses the inductive bias that correlations between location characteristics and population statistics exist, and a learned model is used to capture such correlations to reduce the error. For instance, people tend to spend a similar amount of time in similar locations (e.g., gas stations), and the model will be able to aggregate such information across locations to reduce the error across locations. Furthermore, the correlations between location characteristics and population statistics can be learned from locations where accurate estimates are available and extrapolated to locations for which less data is available. This takes advantage of the fact that more data is available for some neighbourhoods, and allows improving the accuracy for neighbourhoods where lack of data would otherwise have led to large errors, thus providing accurate answers across the board. 

Specifically, our contributions are as follows.
\begin{itemize}
    \item We present the first study of location bias and its impact on releasing aggregate statistics using large-scale GPS datasets.
    \item We present, \name{}, a neural network approach that provides accurate estimates of population statistics by taking into account contextual information
    \item Through extensive experiments on commonly used real-world datasets, we show \name{} improves accuracy by up to 3 times for under-represented neighbourhoods and 2 times across all neighbourhoods, while, surprisingly, statistical debiasing often worsens the accuracy.  
\end{itemize}

The rest of this paper is organized as follows. Sec~\ref{sec:rel_work} discusses related work, Sec.~\ref{sec:setup} describes problem setting, Sec.~\ref{sec:method} discusses our methodology, Sec.~\ref{sec:exp} presents our empirical study and Sec.~\ref{sec:conclusion} concludes the paper. 
\section{Related Work}\label{sec:rel_work}
Bias in observed location data has been documented in a variety of data sources \cite{lu2017understanding, wesolowski2014quantifying, milusheva2021assessing, coston2021leveraging, wesolowski2013impact, wesolowski2013impact, ranjan2012call}. Closest to our setting, \cite{coston2021leveraging} reports similar biases on Safegraph data, showing that older and non-white populations
are less likely to be captured by mobility data. Their analysis compares aggregate statistics released by Safegraph to voter turnout records to uncover biases. Although different in methodology and data sources (they use voter turnout records as population ground truth while we utilize census data), their results corroborate our observations regarding the existing biases in the dataset. Nonetheless, we note that Safegraph does provide an analysis of bias in their datasets \cite{safegraphbias}, where they argue that on aggregate and across the US, the proportion of the observed population in certain demographics across the US is the same as the proportion of the true population in that demographic across the US. As also noted by \cite{coston2021leveraging}, this observation does not imply that there is no sampling bias, because biases at lower granularity can still exist. This is shown in our experiments, where we observe that sampling ratio of census tracts is negatively correlated with their median income. We also observe that sampling ratios are correlated with age groups (similar observation is reported by \cite{coston2021leveraging}), a demographic attribute not considered by Safegraph \cite{safegraphbias} in their analysis. Overall, we have focused on Safegraph dataset because we have access to the raw data (unaggregated GPS signals, see Sec.~\ref{sec:exp}) used for calculating aggregate statistics, but we expect bias in location data to be pervasive across the board and in any dataset collected through mobile apps.

Finally, to the best of our knowledge, there is no existing work that addresses bias in location data, with the related work being confined to only documenting such biases. Although many companies have released such aggregate statistics,  (e.g., Google Mobility Report \cite{googleMobility}, Facebook Social Connectedness Index \cite{facebookSCI} in addition to Safegraph Patterns \cite{sgp}), the statistics are reported by simply aggregating the observed data without accounting for the bias. Related to our setting is the work on synthetic trajectory generation such as \cite{yu2017seqgan, ouyang2018non}, which can be used to generate more data to increase the size of the observed data. However, such approaches will not be able to account for the observed bias, and will merely replicate such biases while creating more data. Finally, our use of neural networks for estimating population statistics follows the NeuroDB framework~\cite{zeighami2023neural, zeighami2023neurodb} and is similar to \cite{zeighami2022neural, ahuja2023neural} and \cite{zeighami2023neural} that, respectively, do so to answer queries in a privacy-preserving manner or for incomplete relational databases.   
\vspace{-0.3cm}
\section{Problem Setup}\label{sec:setup}
\vspace{-0.2cm}
\textbf{Setup}. We are given a dataset of user stay-point sequences, $D$, of $n$ users, where each user's sequence describes locations the user has stayed at in a city.  This stay-point sequence is derived from user trajectories (i.e., raw GPS readings) by extracting the locations at which the users stayed for a long enough duration (we  describe the exact methodology in Sec.~\ref{sec:exp}). For a user $u$, their stay-point sequence is a sequence of $k_u$ stay-points, $<p_1, ..., p_{k_u}>$, for an integer $k_u$. A stay-point, $p$, is a tuple $p=$(\texttt{lat}, \texttt{lon}, \texttt{arrive\_t}, \texttt{leave\_t}), where \texttt{lat} and \texttt{lon} denote the latitude and longitude  where the user stayed, \texttt{arrive\_t} is the time the user entered the location and \texttt{leave\_t} the time the user left the location. We assume $D$ is a subset of a set $\mathcal{D}$, where $\mathcal{D}$ is the set of stay-point sequences of the entire population of size $N$. That is, $\mathcal{D}$ is the set of stay-point sequences of all the population of a city, whereas $D$ is the set of stay-point sequences of the subset of the population whose location has been collected. 

\textbf{Sampling Bias}. We use observed population statistics (from our dataset) and government-released population statistics (from censuses) for the city to study the sampling procedure. 

For each user, we consider their \textit{home neighbourhood} to be the neighbourhood where they spend the most time (i.e., we assume users spend the most time at their home, since they stay there overnight and for long periods). This information is obtained from their list of stay-points. We use the users' home neighbourhoods to compare the observed population of each neighbourhood with its true population. Let $h(u)$ be the home neighbourhood of a user $u$. Then, the observed population, $n_\mu$, of a neighbourhood $\mu$ is $n_\mu=|\{u\in D, h(u)=\mu\}|$.

We consider the government-released statistics for the city to be the true population statistics, i.e., calculated from $\mathcal{D}$. In the US, where our datasets are collected (See Sec.~\ref{sec:exp} for dataset details), US Census releases such information. We utilize aggregate population statistics (e.g., population, median income) released for different census tracts, where census tracts are small (with about 4,000 inhabitants \cite{censustractsize}) subdivisions of counties (similar to zip codes). We consider each census tract as a neighbourhood and use the terms census tract and neighbourhood interchangeably. Let $N_\mu$ be the true population of the neighbourhood $\mu$ (obtained from Census). 

We define the \textit{sampling ratio}, $s_\mu$, of $\mu$ as $s_\mu=\frac{n_\mu}{N_\mu}$, which denotes the fraction of the population from the neighbourhood that is sampled. We consider the setting where the sampling ratio, $s_\mu$, is different for different neighbourhoods $\mu$, the setting we call biased sampling. In this case, for different neighbourhoods, a different portion of their population has been sampled. Biased sampling based on other attributes (e.g., income, race, age) often translates to bias based on neighbourhood, since neighbourhoods in the US are often segregated and homogeneous \cite{homogonousb, homogonousa}. Biased sampling happens often in practice; Table~\ref{tab:sampling_bias} presents one such instance where it shows the dataset used by Safegraph (used by \cite{chang2021mobility,zeighami2021estimating,rambhatla2022toward,hu2021human, haraguchi2022human, song2022factors, zhao2016urban, sun2022predicting,Federal2023CARTS,jin2022customer}%,ballantyne2022framework,safegraph2023clear}) is collected under the biased sampling setting and that the bias is correlated with other demographic attributes. %We use this dataset throughout our experiments. As previously mentioned, a vast body of research and industry work has used Safegraph dataset for analysis purposes \cite{chang2021mobility,zeighami2021estimating,rambhatla2022toward,hu2021human,ihme2021modeling,glaeser2022jue, haraguchi2022human, song2022factors, zhao2016urban, liu2009urban, chang2022role, sun2022predicting,Federal2023CARTS,jin2022customer,ballantyne2022framework,safegraph2023clear}. Thus, it is of immense significance to mitigate the consequences of this biased sampling when releasing population statistics. % commonly used real-world dataset arethat  where we show sampling ratio across neighbourhoods is correlated with the demographics of the neighbourhoods. 

\textbf{Problem Definition}. Our goal is to provide an accurate estimate of population statistics given access only to a biased subsample, that is to provide estimates of a statistic using $D$ so that the answer is similar to answers from $\mathcal{D}$. Population statistics are aggregate queries over $\mathcal{D}$. We specifically focus on aggregate queries over neighbourhoods, and we use the terms \textit{queries} and \textit{statistics} interchangeably. A query $q=(AGG, \alpha, \mu)$ asks for aggregation, $AGG$, of an attribute, $\alpha$, of all the stay-points, $p$, that fall in neighbourhood $\mu$, so that the true answer to a query $q$ is $AGG(\{p[\alpha]|p\in u_{\mu}, u\in \mathcal{D}\})$, where for a user $u$, we denote their set of stay-points in $\mu$ by $u_\mu$, i.e., $u_\mu=\{p, (p[\text{lat}], p[\text{lon}])\in \mu\}$. Although our approach is  generic, we consider \texttt{COUNT} and \texttt{AVG} aggregation functions and specifically queries of \textit{total number of visits}, \textit{average visit duration} and \textit{average distance travelled}. %, we focus on visit duration and distance travelled attributes. 
All such statistics are important for disease spread analysis and policy-making, as well as transportation and urban planning, and Safegraph is already releasing aggregate information for these attributes for different neighbourhoods \cite{sgp}. Specifically, the visit duration of each stay-point $p$ is calculated as $p[$\texttt{leave\_t}$]-$p[\texttt{arrive\_t}$]$. Furthermore, for the $i$-th stay point of a user, $p_i$, the distance travelled is \texttt{dist}(($p_{i}$[\texttt{lat}], $p_{i}$[\texttt{lon}]), ($p_{i+1}$[\texttt{lat}], $p_{i+1}$[\texttt{lon}])), where we use Euclidean distance (since distances are short, there is no need to take curvature of the earth into consideration using geographic coordinate system (GCS)) as our distance function (distance travelled for the user's first stay-point is undefined and ignored in computations). We consider both visit duration and distance travelled to be attributes of the stay-point. The average visit duration and average distance travelled queries ask for the average of visit duration and distance travelled of all the stay-points (across all users) that fall in a neighbourhood, and total number of visits is similarly defined. For a query $q=(AGG, \alpha, \mu)$, we denote by $c_\mu$ the true answer to $q$ if $AGG=$\texttt{COUNT} and by $y_\mu^\alpha$ if if $AGG=$\texttt{AVG}.

\textbf{Sampling Assumption and Terminology}. Although the following sampling assumptions are not required by our methodology, we use these assumptions to analyze and evaluate different approaches. We assume that the data $D$ is sampled from $\mathcal{D}$ as follows. For every neighbourhood, $\mu$, $n_\mu$ i.i.d and uniform samples from $\{u\in \mathcal{D}, h(u)=\mu\}$ are selected, i.e., each sample $X_i^\mu$, $1\leq i\leq n_\mu$ is equal to one of elements of $\{u\in \mathcal{D}, h(u)=\mu\}$ with probability $\frac{1}{N_\mu}$. This sampling procedure is repeated for all neighbourhoods, where we obtain a different number of samples for different neighbourhoods. In practice, this assumption often holds due to the homogeneity of neighbourhoods \cite{homogonousa, homogonousb}. For instance, if an app that collects data is mostly used by the senior population, then more data will be collected from neighbourhoods with a larger older population. However, within the older population, the sampling can be assumed to be uniform.   

Therefore, the database $D$ is a collection of random variables. Next, for concreteness, we review some terminology. For an estimator $\hat{\theta}$, calculated from $D$, to estimate a population statistic $\theta$, recall that bias of $\hat{\theta}$ is $\texttt{bias}(\hat{\theta})=E[\hat{\theta}]-\theta$. The estimator $\hat{\theta}$ is called an unbiased estimator of $\theta$ if $E[\hat{\theta}]=\theta$ and is otherwise called a biased estimator. Furthermore, recall that the mean squared error of an estimator can be written as 
\begin{align}\label{eq:mse}
    E[(\theta-\hat{\theta})^2]=\texttt{bias}(\hat{\theta})^2+\texttt{var}(\hat{\theta}).
\end{align}
\if 0
\begin{itemize}
    \item Given a dataset of user trajectories
    \begin{itemize}
        \item For each user we have a sequence of stay points
        \item Some stay points are at POIs
        \item For each user we know their home census tract
    \end{itemize}
    \item The set is a subsample of the total population
    \item Goal is to find average time people spend at a location
    \item Sampling assumption
    \begin{itemize}
        \item For each census tract we sample a different portion of users
        \item users within a census tract are sampled uniformly at random
        \item for each user all their stay points are sampled
    \end{itemize}

\end{itemize}
\fi
\section{Querying Biased Location Data}\label{sec:method}
In this section, we present \name{}, our neural network approach for answering queries on biased location data. We first discuss the downsides of answering queries oblivious to the bias (Sec.~\ref{sec:oblivious}) or using statistical debiasing (Sec.~\ref{sec:debiased}). Subsequently, in Sec.~\ref{sec:learned} we present our learned approach that addresses such shortcomings.

\subsection{The Oblivious Method}\label{sec:oblivious}
A naive approach to solving the problem is answering queries without considering the bias. 

\textbf{\texttt{AVG}}. For \texttt{AVG} queries, this means reporting answers directly on the observed dataset. Specifically, for a query on attribute $\alpha$ in neighbourhood $\mu$ the estimate from $D$ is
$$
\hat{y}_\mu^\alpha=\frac{\sum_{u\in D}\sum_{p\in u_\mu}p[\alpha]}{\sum_{u\in D}|u_\mu|},
$$
Where to calculate the average visit duration for a neighborhood, we go over all the observed visits in the neighborhood and report their average value as the answer. 

\textbf{\texttt{COUNT}}. For \texttt{COUNT} queries, we need to scale the query answers observed on the sample dataset. Ignoring the sampling bias, we scale the observed answers by $\frac{N}{n}$ to obtain the estimate 
$$
\hat{c}_\mu=\frac{N}{n}\sum_{u\in D}|u_\mu|.
$$

\textbf{Shortcomings}. Intuitively, in this approach and for \texttt{COUNT} queries, if we observe 10\% of the population, we scale our estimates by 10. This is problematic, because, e.g., for a neighbourhood that is mostly visited by seniors, and when the senior population is under-sampled (i.e., less than 10\% of the older population is sampled), then scaling by 10 underestimates the number of stay-points for that neighbourhood. A similar example holds for average queries, e.g., if the senior population stays for a shorter duration than the rest of the population in a neighbourhood, then the average calculated based on the observed samples overestimates the true average.

More theoretically, given our sampling assumption it is easy to see that both $\hat{c}_\mu$ and $\hat{y}_\mu^\alpha$ estimators are biased estimators of the true population statistics. Since the error of an estimator can be decomposed into bias and variance terms, this bias can contribute to large errors for such an approach. Note that $\hat{c}_\mu$ and $\hat{y}_\mu^\alpha$ are biased estimators because the sampling procedure is biased and they do not take this bias into account. That is, if the data was sampled uniformly at random from the population, $\hat{c}_\mu$ and $\hat{y}_\mu^\alpha$ would have been unbiased estimators.   
%We note that, for both COUNT and AVG queries, the approach described above returns an unbiased estimate if the data was not biased. However, due to the bias present, the answers are biased.

\subsection{Statistical Debiasing}\label{sec:debiased}
To reduce the error, our second attempt uses statistical debiasing to provide unbiased estimators of the population statistics, weighing observed samples with their probability. %First consider the count query at location $l$, asking for the total number of visits in $l$. 

\textbf{\texttt{COUNT}}. Consider the users with home neighbourhood $\eta$. %Assume the true population of $\eta$ is $N_\eta$ and its observed population is $n_\eta$. 
To answer queries, we weight the visits of each user from $\eta$ by $\frac{N_\eta}{n_\eta}$. Intuitively, scaling by $\frac{N_\eta}{n_\eta}$ is similar to assuming for every observed user from $\eta$ there are $\frac{N_\eta}{n_\eta}$ (unobserved) users in $\eta$ that have the same characteristics. For instance, if the sampling rate for the seniors is 1\% while the sampling rate for the rest of the population is 10\%, to know how many people are at a location, one needs to scale every observation from seniors by 100 and observations from the rest of the population by 10. Scaling by a larger value helps account for the fact that the older population was under-sampled. 

Let $H$ be the set of all neighbourhoods. Formally, our estimate of the number of people in a neighbourhood $\mu$ is
%is Let $u_\mu$ be the subset of the trajectory of $u$ that falls inside $l$, and let $|u_\mu|$ be the number of elements in that trajectory. Our estimate 
$$\hat{c}_\mu=\sum_{\eta\in H}\frac{N_\eta}{n_\eta}\sum_{u\in D, h(u)=\eta}|u_\mu|.$$

\begin{lemma}
    $\hat{c}_\mu$ is an unbiased estimator of the population statistic $c_\mu$ under the sampling assumptions of Sec.~\ref{sec:setup}.
\end{lemma}
%\vspace{-0.5cm}%Note that the population statistic is $c_\mu=\sum_{h\in H}\sum_{u\in D, h(u)=\eta}|u_\mu|$. It is easy to see that $\hat{c}_\mu$ is an unbiased estimator of $c_\mu$, that is, $E[\hat{c}_\mu]=c_\mu$. Specifically, 
\textit{Proof}.

\begin{align*}
    E[\hat{c}_\mu]&=\sum_{\eta\in H}\frac{N_\eta}{n_\eta}\sum_{u\in D, h(u)=\eta}E[|u_\mu|]
    =\sum_{\eta\in H}\frac{N_\eta}{n_\eta}n_\eta E[|u_\mu|]\\
    &=\sum_{\eta\in H}\frac{N_\eta}{n_\eta}n_\eta\sum_{u_\mu\in \mathcal{D}, h(u)=\eta}\frac{1}{N_\eta}|u_\mu|=c_\mu
\end{align*}\hfill$\square$

\textbf{\texttt{AVG}}. For average queries, obtaining an unbiased estimator is more difficult. For an attribute $\alpha$, let the true attribute sum for neighbourhood $\mu$ be $$t_\mu^\alpha=\sum_{\eta\in H}\sum_{u\in \mathcal{D}, h(u)=\eta}\sum_{p\in u_\mu}p[\alpha].$$ The average of the attribute at $\mu$ is therefore $y_\mu^\alpha=\frac{t_\mu^\alpha}{c_\mu}$. The difficulty in obtaining an unbiased estimator is due to having to estimate both the numerator and the denominator of this quantity. To simplify the discussion, we assume $c_\mu$ is known. To obtain an estimator, we only need to estimate $t_\mu^\alpha$, which can be done by weighting user stay-points similar to \texttt{COUNT}. Specifically, let 

$$\hat{t}_\mu^\alpha=\sum_{\eta\in H}\frac{N_\eta}{n_\eta}\sum_{u\in D, h(u)=\eta}\sum_{p\in u_\mu}p[\alpha].$$ 

Similar to the above, we have that 

\begin{lemma}
    $\frac{\hat{t}_\mu}{c_\mu}=\hat{y}_\mu^\alpha$ is an unbiased estimator of $y_\mu^\alpha$ under the sampling assumptions of Sec.~\ref{sec:setup}.
\end{lemma}
\textit{Proof}.
\begin{align*}
    E[\hat{t}_\mu]&=\sum_{\eta\in H}\frac{N_\eta}{n_\eta}\sum_{u\in D, h(u)=\eta}E[\sum_{p\in u_\mu}p[\alpha]]\\
    &=\sum_{\eta\in H}\frac{N_\eta}{n_\eta}n_\eta E[\sum_{p\in u_\mu}p[\alpha]]
\end{align*}
\begin{align*}
    &=\sum_{\eta\in H}\frac{N_\eta}{n_\eta}n_\eta\sum_{u_\mu\in \mathcal{D}, h(u)=\eta}\frac{1}{N_\eta}\sum_{p\in u_\mu}p[\alpha]\\
    &=t_\mu.
\end{align*}

Therefore, $E[\frac{\hat{t}_\mu^\alpha}{c_\mu}]=y_\mu^\alpha$. \hfill$\square$

In practice, we observed that even with the assumption that $c_\mu$ is known, this unbiased estimator performs poorly, so we do not further relax this assumption. Nonetheless, we note that the estimator $\frac{\hat{t}_\mu}{\hat{c_\mu}}$ (i.e., using our estimate $\hat{c}_\mu$ of $c_\mu$ to estimate the denominator of $a_\mu$) is not an unbiased estimator. %, as proved in Appendix~\ref{appx:proofs}. 

\textbf{Shortcomings}. This approach eliminates bias in query answering, so that the remaining error is due to the variance of the estimators (recall that error can be decomposed into bias and variance, see Eq.~\ref{eq:mse}). Although this helps improve the accuracy for \texttt{COUNT} queries, in practice, we observed that, in the case of \texttt{AVG} queries, the large variance of the unbiased estimator leads to a larger error than the biased estimator. The difference in the effectiveness of debiasing for \texttt{COUNT} and \texttt{AVG} queries can be attributed to the difference in the estimator's variance.  In our dataset, the number of visits of individuals has a much lower variance than the time people stay in different locations or their average distance traveled, as shown in Sec.~\ref{sec:exp:variance}. Thus, debiasing does not help reduce error for average queries, as it does not reduce the variance which is the main source of error. Furthermore, the weights used for debiasing can often be large, further increasing the variance of the debiased estimator, leading to worse accuracy.

\vspace{-0.2cm}
\subsection{Learned Estimation}\label{sec:learned}
\vspace{-0.2cm}
For average queries, the two approaches discussed so far show an example of bias/variance trade-off in estimation, where we see lowering the bias in our estimation increases the variance and leads to worse error. This shows that eliminating bias in our estimator can lead to worse results. Instead, we see that using an estimator with a correct inductive bias is able to provide more accurate results.

To provide lower error, we use a learned estimator to answer queries, where we train a model that uses information about a neighbourhood to estimate query answers. Intuitively, we use the inductive bias that there exist correlations between neighbourhood characteristics and query answers for the neighbourhood, and to reduce the error, a learned model is used to capture such correlations. %, without overfitting to the error in observed values for each neighbourhood. 
For instance, similar POIs tend to have similar visit durations, and the model can aggregate such information across neighbourhoods to reduce error observed for each neighbourhood. Furthermore, such correlations between location characteristics and query answers can be learned from queries for which accurate answers are available. This takes advantage having more data available for some neighbourhoods. Therefore a model can learn accurate patterns from those neighbourhoods that can be extrapolated to neighbourhoods for which less data is available. We provide an overview of this approach in Sec.~\ref{sec:learned:overview} and describe further details in Secs.~\ref{sec:learned:embedding} and \ref{sec:learned:model}.

%Intuitively, our approach can be seen as a variance reduction technique that utilizes information about neighbourhoods. 

\begin{figure}
    \centering
    \includegraphics[width=\columnwidth]{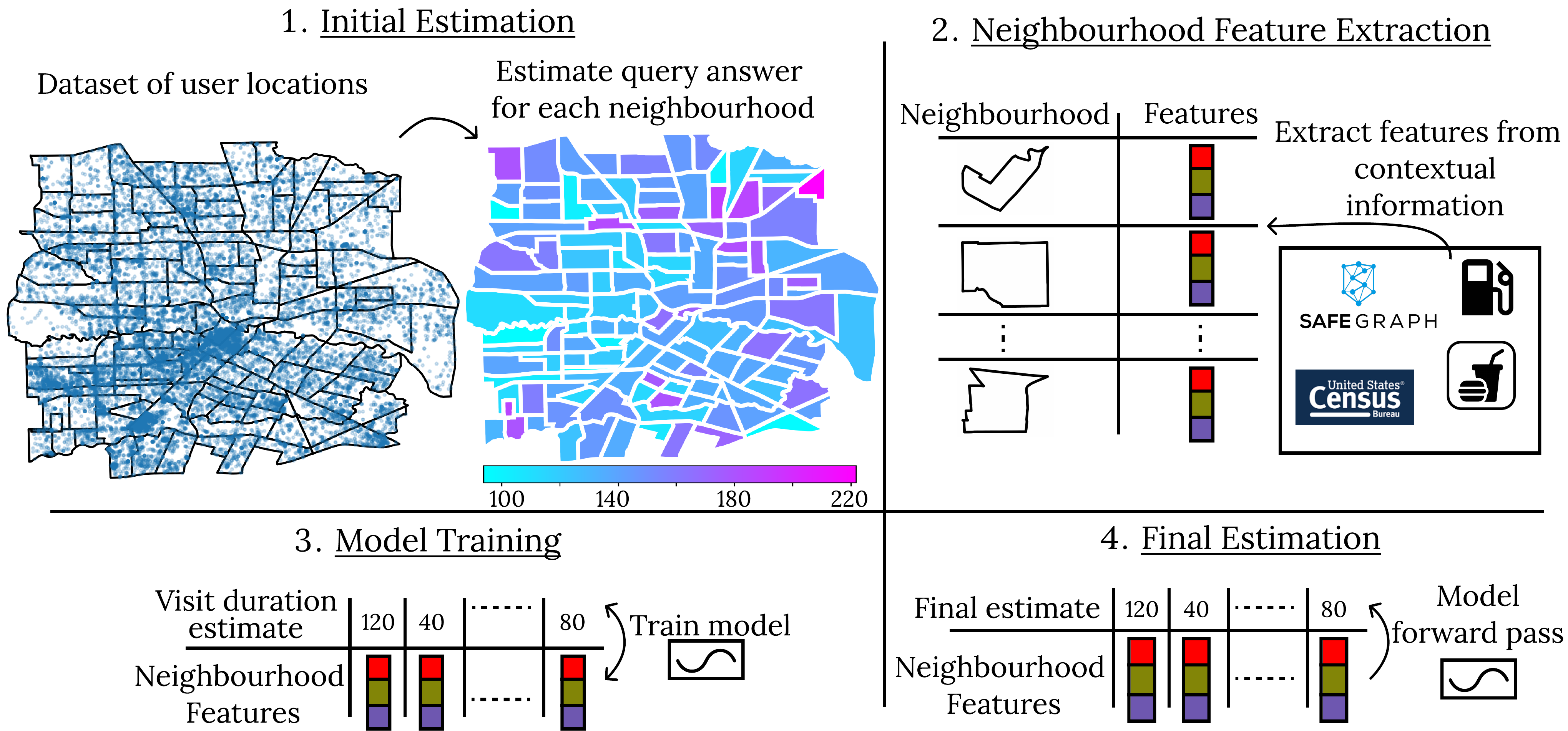}
    \caption{\name{} Overview}
    \label{fig:overview}
\end{figure}

\subsubsection{Overview}\label{sec:learned:overview}
Figure~\ref{fig:overview} shows an overview of \name{}. \name{} has four steps as described below. %PUT A PSUDEO CODE XXX

\textbf{1. Initial Estimation}. We use either the oblivious method of Sec.~\ref{sec:oblivious} or the method using statistical debiasing of Sec.~\ref{sec:debiased} to obtain an initial estimate for the query answer for each neighbourhood. These initial estimates are later used by the model to obtain the final query answers estimate, where the training process extracts the underlying correlations from these initial estimates without overfitting to their error.

\textbf{2. Neighbourhood Feature Extraction}. We create features for each neighbourhood using contextual information available about the neighbourhood through auxiliary data sources. Intuitively, the feature vector captures characteristics of the neighbourhood that are relevant to the query answer, so that a model can learn the correlations between such characteristics and query answers. For instance, the average visit duration in a neighbourhood, is expected to be related to the type of POIs that exist in the neighbourhood. Thus, the neighbourhood features will contain information about types of POIs in the neighbourhood. This step is further described in Sec.~\ref{sec:learned:embedding}.

\textbf{3. Model Training}. Model training uses the initial estimates obtained in Step 1 and neighbourhood feature vectors obtained in Step 2 to learn a neural network through a supervised learning approach where the neighbourhood features are the input to the model and the model is trained to estimate query answers. This step is further described in Sec.~\ref{sec:learned:model}.

\textbf{4. Final Estimation}. The final estimates are obtained by performing a forward pass of the model for each neighbourhood. For each neighbourhood, the neighbourhood features from Step 2 is used, further described in Sec.~\ref{sec:learned:model}.

\vspace{-0.1cm}
\subsubsection{Neighbourhood Feature Extraction}\label{sec:learned:embedding}
We extract a set of features for each neighbourhood from auxiliary data sources. %Neighbourhood features used should depend on the query answered, since different queries depend on different characteristics of neighbourhoods. For instance, the number of visits in a neighbourhood may have a much higher dependence on the population of the neighbourhood compared with the average visit duration in a neighborhood, where the latter depends more on the types of POIs in the neighbourhood. 
%In this section, we describe all the features we used. % In practice, we perform a feature selection step to select only the relevant subset of these features for each query. Feature selection is necessary since redundant features easily lead to the model overfitting to the wrong answers. As described in Sec.~\ref{sec:learned:model}, since we only use error-prone estimates as our training labels our training process is especially sensitive, causing the model to easily overfit.

\textit{POI Features}. We utilize the information of POIs within the neighbourhood to characterise the neighbourhood. Specifically, we use the distribution of POI categories within each neighbourhood. We utilize Safegraph Places \cite{sgplaces} (Safegraph Places \cite{sgplaces} only provides a list of POIs, and is not generated based on user cellphone data), which contains list of  POIs in all neighbourhoods, and for each POI the category it belongs to (e.g., if it's a restaurant or a hospital). The distribution of POI categories is a vector of length $k$, where $k$ is the number of categories, whose $i$-th element is calculated by counting the number of POIs in the $i$-th category in the neighbourhood and dividing it by the total number of POIs in the neighbourhood. 

\textit{Demographic features}. We also use demographic features for each neighbourhood, specifically population and median income of the neighbourhood, both obtained from census.     

%The features we used for each neighbourhood are XXXXXX. We observed that for count queries X are more important while for avg X was more important. We show an ablation study in Sec X

\subsubsection{Model Training and Inference}\label{sec:learned:model}
For a neighbourhood $\mu$, let the query answer estimate obtained in Step 1  be $\hat{y}_\mu$ and let the features obtained in Step 2 be $e_\mu$. 

\textbf{Training}. The training set for our model is $T=\{(e_\mu, \hat{y}_\mu), \forall\mu\}$. We consider two variations of the training process. We use (1) unweighted loss, where the model, $\hat{f}$ is simply trained to predict $\hat{y}_\mu$, that is, to minimize $\sum_{(e_\mu, \hat{y}_\mu)\in T}(\hat{f}(e_\mu;\theta)-\hat{y}_\mu)^2$. Furthremore, (2) we train the model while giving more weight to neighbourhoods for which we are more confident about the estimate. Specifically, let $s_\mu$ be the number of samples observed in a neighbourhood $\mu$ and let $w_\mu=\frac{s_\mu}{\max_\mu s_\mu}$. We use the weighted loss function $\sum_{(e_\mu, \hat{y}_\mu)\in T}w_\mu(\hat{f}(e_\mu;\theta)-\hat{y}_\mu)^2$. Intuitively, using a weighted loss function, the model is trained to capture correlations from neighbourhoods where there are more observations and therefore our initial estimate is more accurate. We use fully connected neural networks. 

Note that the training process uses labels $\hat{y}_\mu$ that are not the ground-truth answers to the queries for a neighbourhood, but instead, an initial estimate obtained from the observed database. The goal of the training process is to learn the underlying correlations of these query answers with respect to the neighbourhood characteristics, without overfitting to the error in the estimated answers. To ensure this, we use small models and early stopping for our training process. That is, we stop the training process before the model fully fits to the training data, as fully fitting to the training data means the model will have the same error as the initial estimate.  Early stopping helps the model stop when it captures the correlations in the data but before it overfits to the training labels. This is further experimentally explored in Sec.~\ref{sec:exp:training_duration}. %Model size and the number of epochs for training are determined on a validation set.    

\textbf{Inference}. Obtaining the final estimate for a neighbourhood $\mu$, is done by performing the forward pass $\hat{f}(e_\mu;\theta)$, where $e_\mu$ is the feature vector obtained in Step 2. We note that $e_\mu$ was in the training set. However, by training a model that captures query answer patterns without overfitting to the training labels, $\hat{f}(e_\mu;\theta)$ will be a better estimate of the query answer than the training label, $\hat{y}_\mu$. This is facilitated by weighting the training samples as well as early stopping, as explained above.

\section{Empirical Study}\label{sec:exp}

\subsection{Experiment Setup and Dataset Details}\label{sec:exp:details}
Our experiments use the dataset provided to us by Veraset \cite{vs}, a data-as-a-service company that provides user location datasets. This dataset is the underlying dataset used by Safegraph~\cite{vstosg}, to provide aggregate population statistics, while also used by various other entities \cite{vs_usecase2, vs_usecase}, among them the US government \cite{vs_usecase3}. We first describe this dataset and our preprocessing method in detail and then proceed to discuss the evaluation setup.  

\subsubsection{Dataset Details}
We use the dataset provided to us for December 2019. The dataset consists of records of the form \texttt{user\_id}, \texttt{latitude}, \texttt{longitude} and \texttt{timestamp}, where each record is obtained through phone GPS signals. From this dataset, we extract the stay-points sequences for users to obtain the dataset of the form described in Sec.~\ref{sec:setup}. We perform Stay Point Detection (SPD) \cite{ye2009mining} on the data to remove location signals when a person is moving, and to extract POI visits when a user is stationary.

\begin{table}[t!]
\hspace*{-0.3cm}
    \centering
    \begin{tabular}{c|c|c|c}
        \textbf{City} & \begin{tabular}{@{}c@{}}\textbf{Observed} \\ \textbf{Pop.}\end{tabular} &  \begin{tabular}{@{}c@{}}\textbf{Sampling} \\ \textbf{Ratio}\end{tabular} & \textbf{\# Stay-Points}  \\\hline
Houston&94,355 &0.04 & 1,002,389\\\hline    
Chicago&133,178 &0.03 & 1,493,640\\\hline
San Francisco&24,855 &0.03 & 938,500\\\hline
Tulsa&26,976 &0.04 & 277,077\\\hline
Fargo&6,246 &0.04 & 92,029
    \end{tabular}
    \caption{Summary of dataset statistics}
    \label{tab:data_stat}
\end{table}

\begin{figure}
\begin{floatrow}
%\hspace*{-2cm}
\ffigbox{%
    \centering
    %\begin{minipage}{0.49\columnwidth}
        \includegraphics[width=\columnwidth]{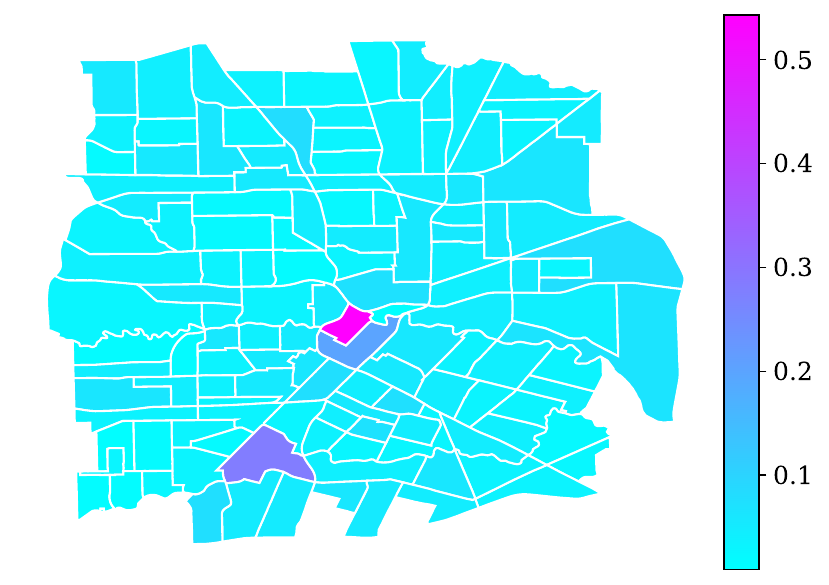}
        }{
        \caption{Sampling Ratios across Houston}
        \label{fig:sampling_ratio_houston}
        }
    %\end{minipage}
    %\hfill
    %\begin{minipage}{0.49\columnwidth}
\ffigbox{%
       \includegraphics[width=\columnwidth]{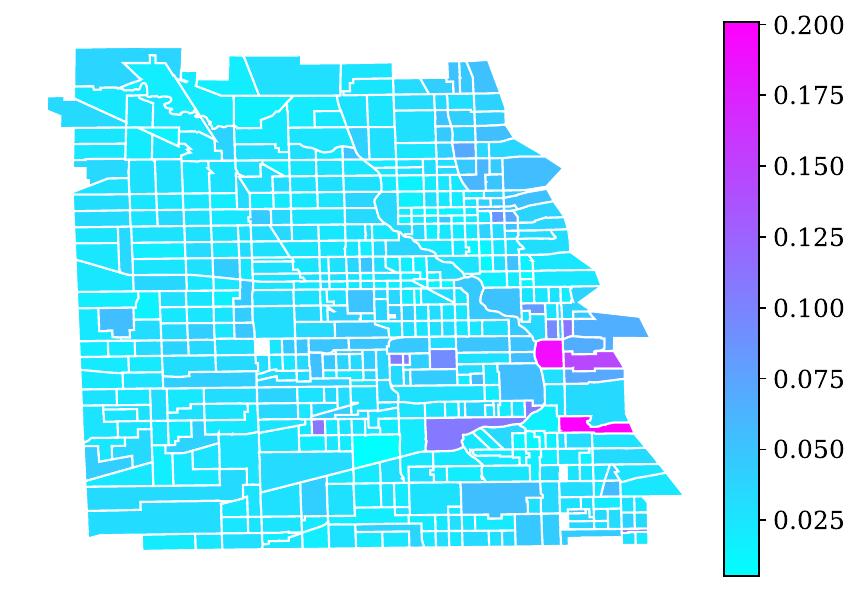}
       }{
        \caption{Sampling Ratios across Chicago}
        \label{fig:sampling_ratio_cook}
        }
    %\end{minipage}
\end{floatrow}
\end{figure}

\begin{figure}[t]
    \centering
    \includegraphics[width=\columnwidth]{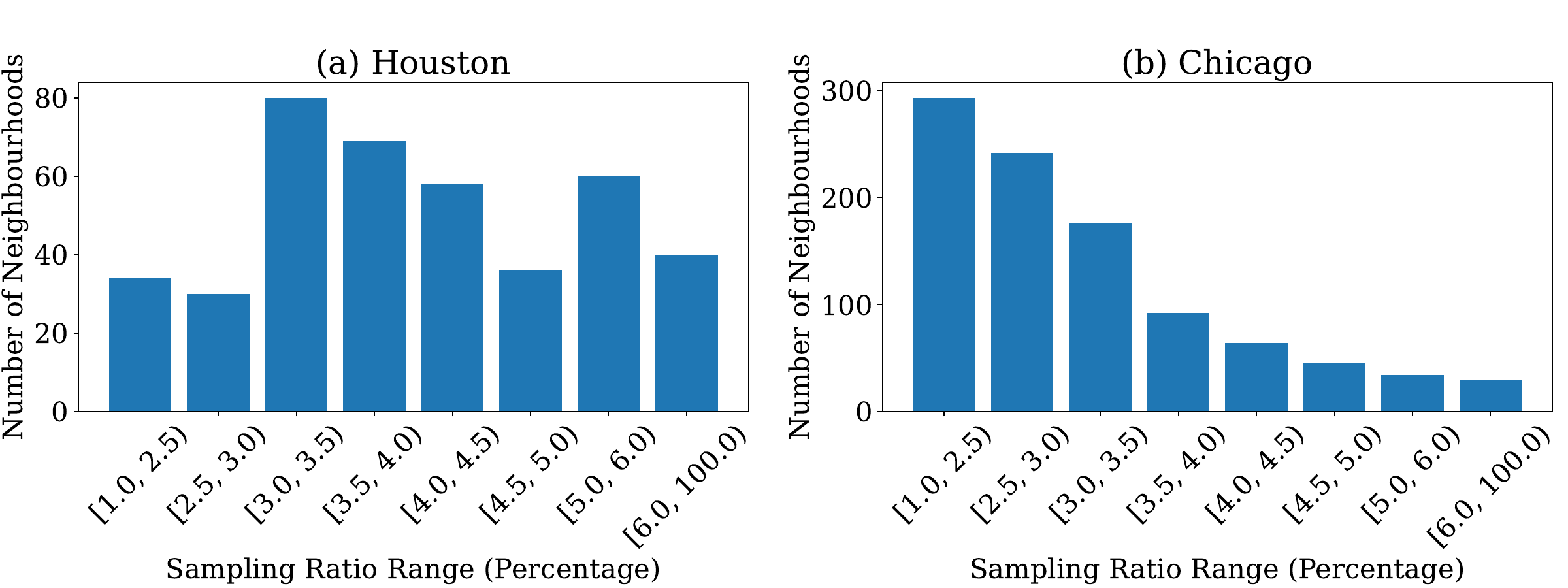}
    \caption{Sampling Ratios Across Neighbourhoods}
    \label{fig:sampling_ratios}
\end{figure}

Table \ref{tab:data_stat} shows the details of our datasets after the above preprocessing steps. The sampling ratios for different cities reported in the third column of the table are calculated by finding the user's home neighbourhood and utilizing its corresponding Census tract demographics as the true population of a city, as described in Sec.~\ref{sec:setup}.
Figs.~\ref{fig:sampling_ratio_houston} and \ref{fig:sampling_ratio_cook} visualize that this sampling ratio is not distributed evenly across different neighbourhoods, and for some neighbourhoods our dataset contains data for a larger proportion of their population. Fig.~\ref{fig:sampling_ratios} further quantifies this, showing how the sampling ratio varies across neighbourhoods.  

To understand factors impacting the sampling ratio, we calculated its correlation between different demographic attributes of each neighbourhood. Specifically, we obtained the median income for each neighbourhood from the US Census, and calculated the Pearson correlation coefficient between the median income of each neighbourhood and its sampling ratio. To calculate the correlation between age attributes, we obtained the population of a neighbourhood in an age range and divide it by the total population of that neighbourhood (both statistics obtained from census). Subsequently, we calculate the correlation coefficient between the calculated normalized age and sampling ratio for each neighbourhood. This normalization by total population is because we are interested in the  correlation with sampling \textit{ratio}, and not the size of the samples. These correlation statistics were reported in Table~\ref{tab:sampling_bias}, showing that, in our dataset, less data is available for seniors or children, while more data is available for the low-income population (we define the child age group as people below 18, adult as 18-65 and senior as above 65). Overall, the amount of data available depends on how the data was collected, the information that Veraset does not publically provide. However, we suspect that more data for low-income population is due to use of Android apps for data collection instead of iPhone apps, which are used more often by the higher-income population. 

\subsubsection{Experiment Setup}\hfill\\
\textbf{Datasets}. To evaluate our methods and to have access to a ground-truth, we consider the Veraset dataset, $D$, to be the true population and we sample a new subset, $D_s$ from this dataset and consider it as the observed dataset. The sampling procedure is designed to mimic the true sampling procedure (based on which $D$ was obtained). Specifically, we use the sampling ratios calculated using $D$ (i.e., sampling ratios obtained by comparing $D$ to the census data) for our sampling procedure. This ensures a similar sampling procedure to what was used to obtain $D$ from the true population, is followed to obtain $D_s$ from $D$. Specifically, for each neighbourhood $\mu$ whose sampling ratio is $\mu_s$ and whose population based on $D$ is $n^D_\mu$, we sample $\mu_s\times n^D_\mu$ users from users in $D$ whose home neighbourhood is $\mu$. This mimics the biased sampling process, where some neighbourhoods have a larger portion of their population observed. We sampled sampling uniformly within each neighbourhood from the population, which, as discussed before, can be true in practice since neighbourhoods themselves are often homogenous \cite{homogonousa, homogonousb}. 

In our experiments, we use datasets corresponding to multiple cities in the US, namely, Houston, Chicago, San Francisco, Tulsa and Fargo. The data for each city is extracted by defining an area of about 20x20 km$^2$ covering the city. For all algorithms, we sample the datasets five times and report the average error across the runs and its standard deviation. 

\textbf{Evaluation Metric}. We evaluate our approach by considering three different estimation tasks, i.e.,  estimating the average visit duration, the average distance traveled and the total number of visits. Each estimate is for a different census tract (i.e., neighbourhood, see Sec.~\ref{sec:setup}) within a city and we report average relative error across the neighborhoods within a city. Specifically, let the true statistic (i.e., calculated from $D$) for a census tract $\mu$ from all tracts $M$  be $x_\mu$, and the estimate obtained from an algorithm (where the algorithm only has access $D_s$) be $\hat{x}_\mu$. We calculate the relative error over all census tracts as $\frac{1}{|M|}\sum_{\mu\in M}\frac{|x_\mu-\hat{x}_\mu|}{|x_\mu|}$. 

Additionally, we also subdivide the census tracts into five categories and report average relative error for each category. The categories are defined based on how much of the data in each census tract was sampled. For each census tract, $\mu$, let $L_\mu$ be the number of stay-points within the census tract and let $l_\mu$ be the number of stay-points sampled within that census tract. Then, we define the \textit{stay-point sampling ratio} as $\frac{l_\mu}{L_\mu}$. Note that this is different from the \textit{user sampling ratio}, defined in Sec.~\ref{sec:setup}, which considers how many of the users belonging to a neighbourhood were sampled. The user sampling ratio is used to describe the bias in sampling, which leads to different values for stay-point sampling ratios across the neighbourhood. On the other hand, stay-point sampling ratio is more correlated with the final error, since it is directly calculated based on the number of stay-point observations in a census tract. To understand the effect of stay-point sampling ratio, we divide the census tracts into five categories based on their stay-point sampling ratio. We consider the quantiles of sampling ratios within a city, and define the five categories as less than first quantiles, between first and second, between second and third, between third and fourth and more than fourth quantiles. 

\begin{table*}[t]
    \centering
    \begin{tabular}{c|c|c|c|c|c|c}
\textbf{City}&\textbf{Oblivious}&\textbf{Debiased}&\textbf{BiasBuster-O}&\textbf{BiasBuster-D}&\textbf{BiasBuster-OW}&\textbf{BiasBuster-DW}\\\hline
Houston&0.41 ($\pm$0.008)&0.43 ($\pm$0.007)&\textbf{0.20} ($\pm$0.007)&\textbf{0.20} ($\pm$0.004)&0.22 ($\pm$0.011)&0.21 ($\pm$0.008)\\\hline
Chicago&0.49 ($\pm$0.013)&0.50 ($\pm$0.016)&0.24 ($\pm$0.008)&0.24 ($\pm$0.009)&\textbf{0.22} ($\pm$0.008)&\textbf{0.22} ($\pm$0.008)\\\hline
San Francisco&0.43 ($\pm$0.028)&0.45 ($\pm$0.026)&0.21 ($\pm$0.019)&0.20 ($\pm$0.019)&\textbf{0.19} ($\pm$0.010)&\textbf{0.19} ($\pm$0.024)\\\hline
Tulsa&0.45 ($\pm$0.020)&0.45 ($\pm$0.030)&0.25 ($\pm$0.026)&0.26 ($\pm$0.033)&0.24 ($\pm$0.027)&\textbf{0.23} ($\pm$0.022)\\\hline
Fargo&0.34 ($\pm$0.032)&0.34 ($\pm$0.053)&0.25 ($\pm$0.027)&0.24 ($\pm$0.045)&\textbf{0.23} ($\pm$0.027)&0.24 ($\pm$0.050)
    \end{tabular}
    \caption{Relative Error for Average Visit Duration}
    \label{tab:res_duration}
\end{table*}
\begin{figure*}[t]
    \centering
    \includegraphics[width=\textwidth]{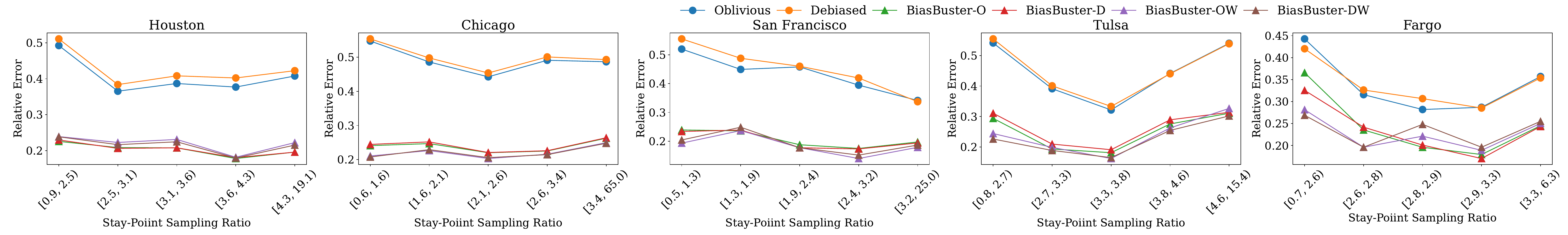}
    \caption{Error Across Different Sampling Ratios for Average Visit Duration Travelled}
    \label{fig:error_breakdown_duration}
\end{figure*}
\begin{table*}[t]
    \centering
    \begin{tabular}{c|c|c|c|c|c|c}
\textbf{City}&\textbf{Oblivious}&\textbf{Debiased}&\textbf{BiasBuster-O}&\textbf{BiasBuster-D}&\textbf{BiasBuster-OW}&\textbf{BiasBuster-DW}\\\hline
Houston&0.49 ($\pm$0.023)&0.48 ($\pm$0.029)&0.41 ($\pm$0.053)&0.36 ($\pm$0.033)&0.27 ($\pm$0.015)&\textbf{0.23} ($\pm$0.012)\\\hline
Chicago&0.73 ($\pm$0.026)&0.74 ($\pm$0.020)&0.44 ($\pm$0.154)&0.52 ($\pm$0.038)&\textbf{0.36} ($\pm$0.028)&0.38 ($\pm$0.032)\\\hline
San Francisco&0.48 ($\pm$0.031)&0.50 ($\pm$0.019)&0.30 ($\pm$0.064)&0.26 ($\pm$0.035)&0.24 ($\pm$0.025)&\textbf{0.23} ($\pm$0.020)\\\hline
Tulsa&0.46 ($\pm$0.070)&0.46 ($\pm$0.072)&0.38 ($\pm$0.037)&0.44 ($\pm$0.047)&\textbf{0.33} ($\pm$0.022)&\textbf{0.33} ($\pm$0.038)\\\hline
Fargo&0.44 ($\pm$0.066)&0.45 ($\pm$0.091)&0.48 ($\pm$0.188)&0.51 ($\pm$0.107)&0.34 ($\pm$0.063)&\textbf{0.33} ($\pm$0.090)
    \end{tabular}
    \caption{Relative Error for Average Distance Travelled}
    \label{tab:res_distance}
\end{table*}

\begin{figure*}[t]
    \centering
    \includegraphics[width=\textwidth]{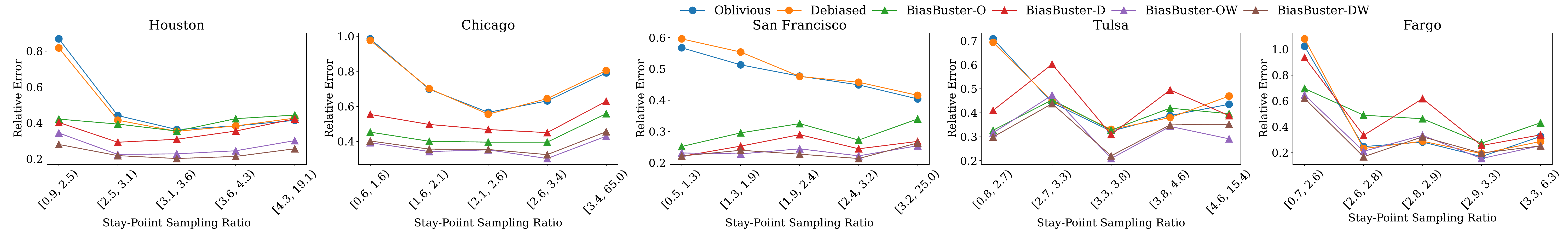}
    \caption{Error Across Different Sampling Ratios for Average Distance Travelled}
    \label{fig:error_breakdown_dist}
\end{figure*}

\textbf{Methods}. We present results for (1) oblivious estimation discussed in Sec.~\ref{sec:oblivious}, referred to as Oblivious, (2) debiased estimation discussed in Sec.~\ref{sec:debiased}, referred to as Debiased and (3) variations of \name{} presented in Sec.~\ref{sec:learned}. Specifically, we train \name{} with labels obtained from both Oblivious and Debiased, respectively referred to as \name{}-O and \name{}-D. Furthermore, we also train both these variations with with weighted loss function which are referred to as \name{}-OW and \name{}-DW. All variations of \name{} are fully connected neural networks with 3 hidden layers and each layer of size 80.

\subsection{Evaluation}\label{sec:exp:eval}
%In this section, we present our experimental results for the three estimation tasks.

\subsubsection{Average visit duration} Table~\ref{tab:res_duration} and Fig.~\ref{fig:error_breakdown_duration} show the results for this task. First consider Table~\ref{tab:res_duration}. It summarizes the error for different cities and different methods where for each column the number in parenthesis is the standard deviation of error across runs. First, we observe that across cities, all variations of \name{} outperform both Oblivious and Debiased significantly, reducing error by more than half in all instances. Second, Debiased performs worse than Oblivious across datasets, indeed showing that debiasing does not help improve accuracy as the large variance in estimation causes large errors (we further discuss variance in estimation in Sec.~\ref{sec:exp:variance}). Finally, we see that all variations of \name{} perform similarly, although the weighted variations, namely \name{}-OW and \name{}-DW often achieve better error (and/or lowest standard deviation), showing a marginal benefit for weighting the samples. We also note that although Debiased consistently performs worse than Oblivious, \name{}-D and \name{}-O (or similarly \name{}-DW and \name{}-OW) have a similar accuracy across datasets, showing that, using \name{}, it is possible to use worse training labels but still achieve good accuracy.

Furthermore, Fig.~\ref{fig:error_breakdown_duration} further breaks down the error for different algorithms. We report the average error for different categories of census tracts defined based on stay-point sampling ratio of the census tracts (as described in Sec.~\ref{sec:exp:details}). Overall, we see that all variations of \name{} provide consistent accuracy across different sampling ratio, while for both Oblivious and Debiased, the error for census tracts with small sampling ratio is often more than 10\% higher than the error for census tracts with higher sampling ratios. This shows that \name{} is able to provide consistent accuracy across neighbourhoods, even when their sampling ratio is small, thus avoiding penalizing communities for which less data is available (e.g., the seniors or children).

\subsubsection{Average distance traveled} Table~\ref{tab:res_distance} and Fig.~\ref{fig:error_breakdown_dist} show the results for the task of estimating average distance traveled. The main observations are similar to the case of average visit duration. Table ~\ref{tab:res_distance} shows that Debiased does not perform better than Oblivious, while \name{} variations significantly outperform both. Moreover, Fig.~\ref{fig:error_breakdown_dist}  shows that \name{} provides consistently low accuracy across different sampling ratios, while both Oblivious and Debiased have very large variations in accuracy across sampling ratios (e.g, in Houston the error drops from 80\% for locations with low sampling ratios to 40\% for locations with high sampling ratios). 

On the other hand, compared with average visit duration, for average distance traveled, we see that weighting the loss function has a more significant impact, where we often see more than 10\% reduction in error when using the weighted loss. Overall, this shows that in the case of distance traveled, the model is able to better extrapolate the patterns from neighbourhoods with a large number of samples to neighbourhoods with a small number of samples. This can be because transferrable patterns from highly-sample to less-sampled neighbourhoodds are more prominent in the case of average distance traveled compared with average visit duration.   

\begin{figure*}[t]
\vspace{-0.4cm}
\begin{floatrow}
\hspace*{-2cm}
\ffigbox{%
    \centering
    \includegraphics[width=0.4\columnwidth]{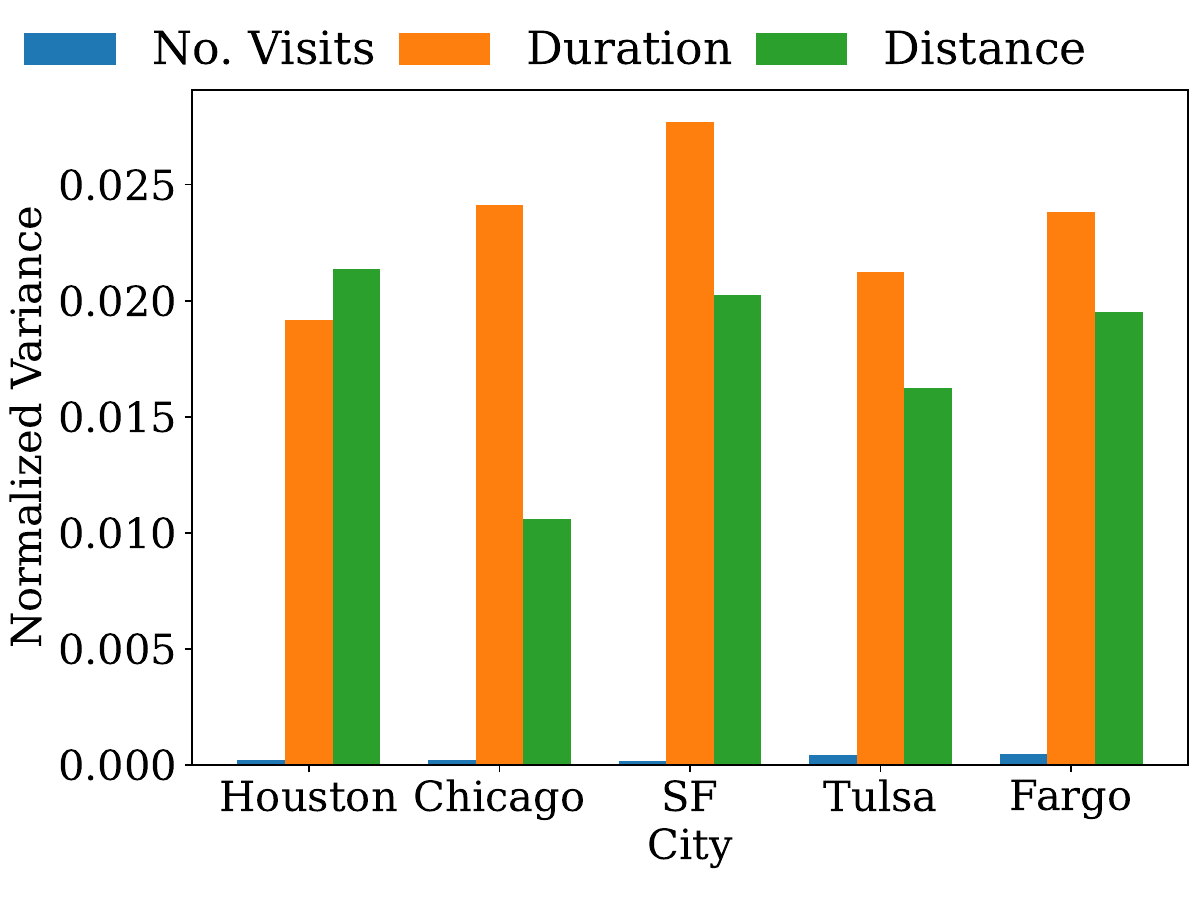}
}{%
    \caption{Variance of different \newline attributes across users}
    \label{fig:variance}
}
\capbtabbox{%
\hspace{-2.5cm}
    \begin{tabular}{c|c|c|c|c}
\textbf{City}&\textbf{Oblivious}&\textbf{Debiased}&\textbf{BiasBuster-D}&\textbf{BiasBuster-DW}\\\hline
Houston&0.70 ($\pm$0.052)&0.46 ($\pm$0.034)&\textbf{0.41} ($\pm$0.039)&0.46 ($\pm$0.072)\\\hline
Chicago&1.07 ($\pm$0.076)&0.63 ($\pm$0.028)&\textbf{0.55} ($\pm$0.022)&0.64 ($\pm$0.049)\\\hline
San Francisco&2.79 ($\pm$0.270)&0.63 ($\pm$0.069)&\textbf{0.59} ($\pm$0.074)&0.63 ($\pm$0.058)\\\hline
Tulsa&0.81 ($\pm$0.043)&0.46 ($\pm$0.071)&\textbf{0.43} ($\pm$0.047)&0.49 ($\pm$0.082)\\\hline
Fargo&0.98 ($\pm$0.208)&\textbf{0.41} ($\pm$0.074)&\textbf{0.41} ($\pm$0.043)&0.44 ($\pm$0.073)
    \end{tabular}
    }
{%
    \caption{Relative Error for Number of Visits}
    \label{tab:res_count}}
\end{floatrow}
\end{figure*}

\if 0
\begin{table*}[t]
    \centering
    \begin{tabular}{c|c|c|c|c}
\textbf{City}&\textbf{Oblivious}&\textbf{Debiased}&\textbf{BiasBuster-D}&\textbf{BiasBuster-DW}\\\hline
Houston&0.70 ($\pm$0.052)&0.46 ($\pm$0.034)&\textbf{0.41} ($\pm$0.039)&0.46 ($\pm$0.072)\\\hline
Chicago&1.07 ($\pm$0.076)&0.63 ($\pm$0.028)&\textbf{0.55} ($\pm$0.022)&0.64 ($\pm$0.049)\\\hline
San Francisco&2.79 ($\pm$0.270)&0.63 ($\pm$0.069)&\textbf{0.59} ($\pm$0.074)&0.63 ($\pm$0.058)\\\hline
Tulsa&0.81 ($\pm$0.043)&0.46 ($\pm$0.071)&\textbf{0.43} ($\pm$0.047)&0.49 ($\pm$0.082)\\\hline
Fargo&0.98 ($\pm$0.208)&\textbf{0.41} ($\pm$0.074)&\textbf{0.41} ($\pm$0.043)&0.44 ($\pm$0.073)
    \end{tabular}
    \caption{Relative Error for Number of Visits}
    \label{tab:res_count}
\end{table*}
\begin{figure}[t]
    \centering
    \includegraphics[width=0.5\columnwidth]{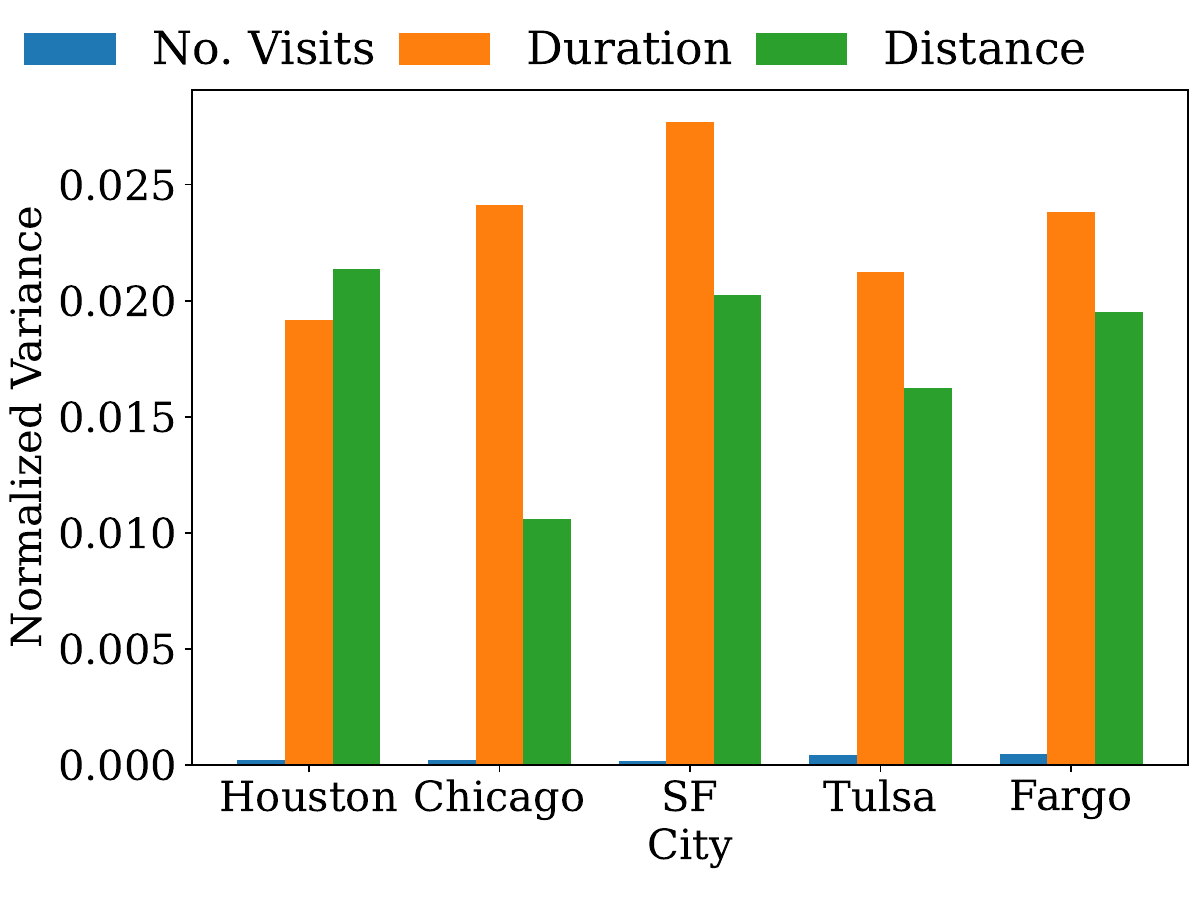}
    \caption{Variance of different attributes across users}
    \label{fig:variance}
\end{figure}
\fi

\begin{figure*}[t]
    \centering
    \includegraphics[width=\textwidth]{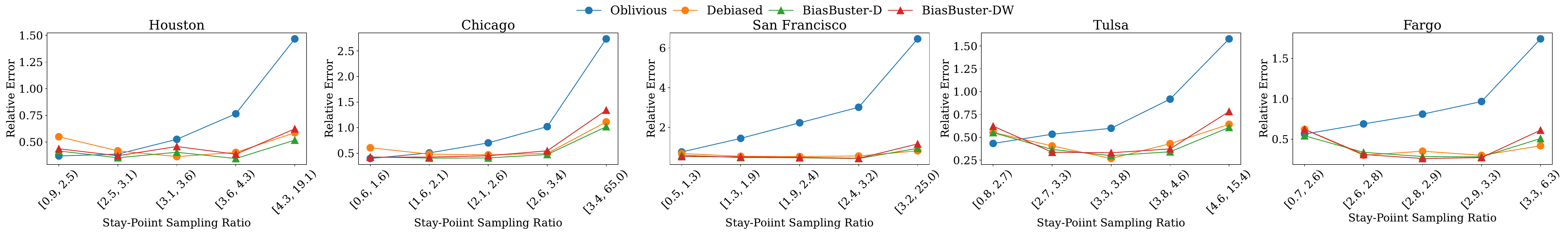}
    \caption{Error Across Different Sampling Ratios for Number of Visits}
    \label{fig:error_breakdown_count}
\end{figure*}

%\vspace{-0.2cm}
\subsubsection{Total Number of Visits} Finally, Table~\ref{tab:res_count} and Fig.~\ref{fig:error_breakdown_count} show the results for the task of estimating the total number of visits. In Table~\ref{tab:res_count}, we see that in contrast to the two average statistics, Debiased does in fact improve upon Oblivious significantly. As shown later (Sec.~\ref{sec:exp:variance}), this is due to the lower variance in number of visits of individuals, compared with average visit duration or distance traveled. Debiasing can multiply estimates by large values, leading to high errors when variance is large, while when the variance is small this operation is effective. 

Due to the significant difference between Oblivious and Debiased, we only train \name{} using Debiased as its labels, so that we only report results for \name{}-D and \name{}-DW. We see that \name{} outperforms Debiased across all cities except Fargo. This can be because for the case of Fargo, \name{} does not have enough training data, since Fargo has the least number of neighbourhoods across all the cities. Finally, we see that \name{}-D performs better than \name{}-DW. Recall that \name{}-DW weights neighbourhoods with more observed samples more heavily. In the case of average queries, this is helpful, as one expects to be able to learn patterns from such neighbourhoods and extrapolate to neighbourhoods with fewer samples. However, for total number of visits, such extrapolation is not as effective. This is because in the case of number of visits, neighbourhoods with more observed samples also tend to have more true visits. Therefore learning from such neighbourhoods leads to overestimating the  number of visits for neighbourhoods where number of observed sampled is small.  

Finally, in Fig.~\ref{fig:error_breakdown_count}, we see that, similar to the other estimation tasks, \name{} is able to provide consistent accuracy across sampling ratios, while the accuracy of Oblivious degrades significantly for neighbourhoods with large sampling ratios.  
Such a degradation is expected for Oblivious, because it scales the answer for all neighbourhoods with a fixed constant $x$. However, neighbourhoods with higher sampling ratios have more than $\frac{1}{x}$ of their data observed, and therefore should be scaled with a smaller scaling factor. This leads to overestimation for such neighbourhoods resulting in a large error.

\vspace{-0.3cm}
\subsection{Variance Analysis}\label{sec:exp:variance}
\vspace{-0.2cm}
In Sec.~\ref{sec:exp:eval}, we saw that debiasing, compared with the oblivious approach, fails to improve accuracy in the case of \texttt{AVG} queries, while it does improve accuracy for the \texttt{COUNT} query. Here, we provide more explanation for this observation.

Recall that error of an estimator can be decomposed into its bias and variance (Eq.~\ref{eq:mse}). For our estimators (both Oblivious and Debiased), their variance depends on the variance of the attribute estimated. For instance, if the entire population has exactly 1 visit to a neighbourhood, then the variance of both Oblivious and Debiased methods for estimating number of visits will be zero (independent of the observed sample), because any sample will only contain users with exactly 1 visit. However, the variance becomes larger if different people have different number of visits (e.g., some have 100 visits while others have only 1). As such, variance of the estimated attribute is an important factor contributing to the final error.

In Fig.~\ref{fig:variance}, we plot the variance of the three attributes studied across different cities. To calculate the variance for the number of visits, we (1) for each neighbourhood calculate how many visits each user has in that neighbourhood, (2) calculate the variance of the number of visits across users for each neighbourhood and (3) present the average variance across all the neighbourhoods. For average visit duration and distance travelled, the process is the same, but in step (1), instead of calculating the number of visits of each user per neighbourhood, we calculate the average visit duration or distance travelled for the user in each neighbourhood. To be able to compare across the attributes, all the values are normalized to be in [0, 1] in step (2) and before calculating the variance, by deducting the minimum from all attributes and dividing them by the maximum value across the users.

Fig.~\ref{fig:variance} shows that both average visit duration and average distance traveled have a larger variance compared with the number of visits. This implies that the error in Oblivious for the average queries is more likely to be due to their variance, and not the bias; on the other hand the variance for the number of visits is very small (smaller than the average queries by orders of magnitude). Since debiasing only reduces the bias and not the variance, it does not help improve the accuracy for average queries (where variance is large), but it does improve accuracy for count queries (where variance is small).   

%Putting this result in the context of Sec.~\ref{sec:exp:eval}, we conclude that \name{} improves the accuracy more significantly in cases where the variance of the attribute estimated is large. 

\begin{figure}[t]
    \centering
    \includegraphics[width=\columnwidth]{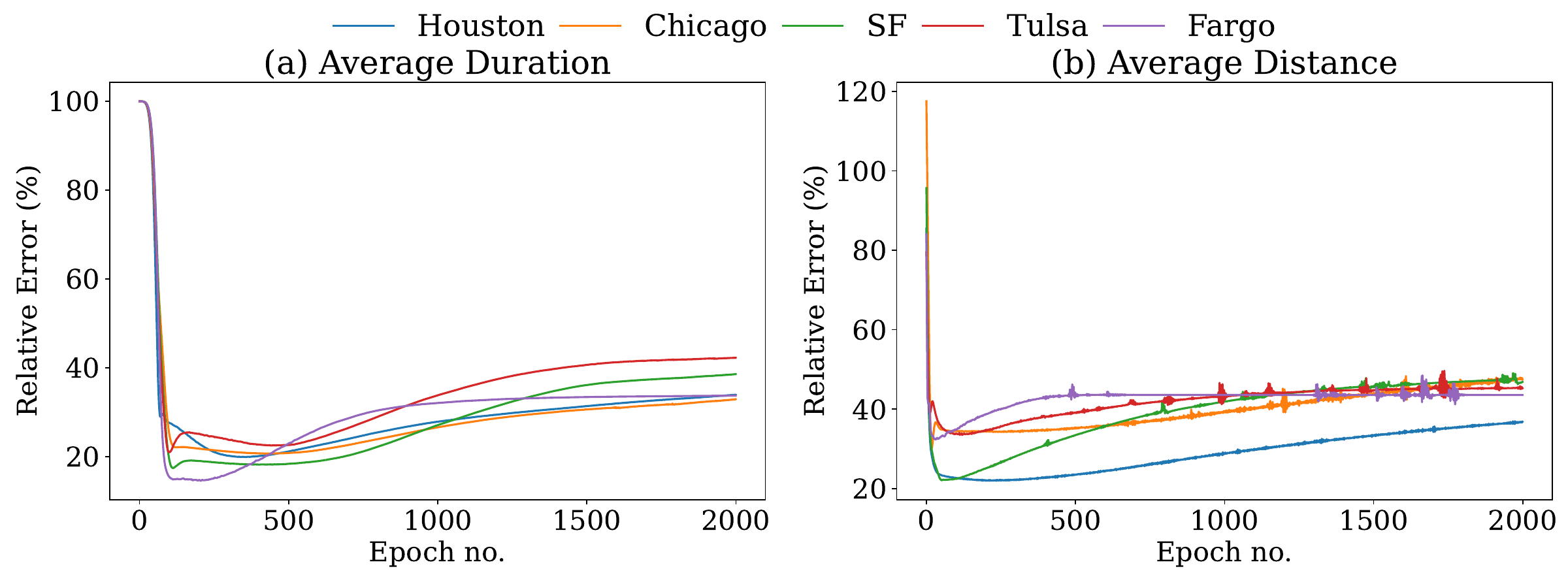}
    \caption{Test accuracy during training processing}
    \label{fig:train_epochs}
\end{figure}

\vspace{-0.2cm}
\subsection{Impact of Training Duration}\label{sec:exp:training_duration}
Since the training process uses labels from the observed data (that is, the training labels are inaccurate), the training process is prone to overfitting. We show this in Fig.~\ref{fig:train_epochs}, where it depicts test accuracy (i.e., accuracy with respect to the true labels) at different stages of training. We see that throughout training, the test accuracy initially improves but worsens after 500 epochs, across cities and for both attributes estimated, although the models' best accuracy is achieved sooner for the average distance travelled. This shows that early stopping (i.e., stopping before the model fully fits the training set) is important for training accurate models.

\if 0
- Show how error changes based on different features

- Other system parameters? 

\begin{itemize}
    \item Dataset and Bias analysis
    \item Setup 
    \begin{itemize}
        \item maybe have some of the stuff from the problem setup  here
    \end{itemize}
    \item Exp with error across cities and sampling factors
    \item ablation study of different features
    \item Show final error correlation to sensitive attributes?
\end{itemize}

\fi
\vspace{-0.2cm}\section{Conclusion}\label{sec:conclusion}\vspace{-0.2cm}
We presented the first comprehensive analysis of bias in real-world location data collected from mobile phones. We showed that this bias exists in a commonly used location dataset (used by Safegraph), that is often utilized for, among other tasks, the sensitive task of COVID forecasting and policy making. We showed that a statistical debiasing method often fails to improve accuracy, and instead presented \name{}, a neural network approach that utilizes correlations between location characteristics and population statistics to provide accurate estimates of population statistics. Our experiments showed that \name{} improves accuracy by up to 3 times for underrepresented populations. Future work includes extending location feature extraction to further improve accuracy and considering other types of aggregate queries.% and incorporating the \name{} estimates into downstream tasks such as COVID-19 forecasting.   

\bibliographystyle{IEEEtran}
\bibliography{references}
\end{document}